\documentclass[11pt]{article}

\usepackage[preprint]{acl}

\usepackage{times}
\usepackage{latexsym}

\usepackage[T1]{fontenc}

\usepackage[utf8]{inputenc}

\usepackage{microtype}

\usepackage{inconsolata}

\usepackage{graphicx}
\usepackage{booktabs}
\usepackage{multirow}
\usepackage{amsmath}
\usepackage{amssymb}
\usepackage{bm}
\usepackage{algorithm}
\usepackage{algorithmic}
\usepackage{placeins}
\usepackage[table]{xcolor}
\usepackage[most]{tcolorbox}
\newcommand{\std}[1]{{\fontsize{6pt}{6pt}\selectfont\,$\pm$\,#1}}

\definecolor{gskillpromptbg}{HTML}{EFF6FF}
\definecolor{gskillpromptframe}{HTML}{2E87EE}
\definecolor{gskilltaskbg}{HTML}{FFF3E8}
\definecolor{gskilltaskframe}{HTML}{F56301}
\definecolor{gskillstepbg}{HTML}{F5EEFF}
\definecolor{gskillstepframe}{HTML}{6801CA}
\definecolor{gskillslot}{HTML}{9A3412}
\newtcolorbox{gskillpromptbox}[1]{
  enhanced,breakable=false,colback=gskillpromptbg,colframe=gskillpromptframe,
  colbacktitle=gskillpromptframe,coltitle=white,fonttitle=\small\bfseries,
  title={#1},boxrule=0.5pt,arc=1.5mm,left=1.5mm,right=1.5mm,
  top=1mm,bottom=1mm,before skip=5pt,after skip=5pt}
\newtcolorbox{gskilltaskbox}[1]{
  enhanced,breakable=false,colback=gskilltaskbg,colframe=gskilltaskframe,
  colbacktitle=gskilltaskframe,coltitle=white,fonttitle=\small\bfseries,
  title={#1},boxrule=0.5pt,arc=1.5mm,left=1.5mm,right=1.5mm,
  top=1mm,bottom=1mm,before skip=5pt,after skip=5pt}
\newtcolorbox{gskillstepbox}[1]{
  enhanced,breakable=false,colback=gskillstepbg,colframe=gskillstepframe,
  colbacktitle=gskillstepframe,coltitle=white,fonttitle=\small\bfseries,
  title={#1},boxrule=0.5pt,arc=1.5mm,left=1.5mm,right=1.5mm,
  top=1mm,bottom=1mm,before skip=5pt,after skip=5pt}

\title{G-Skill: Graph-Grounded Skill Evolution\\
for Agentic Reinforcement Learning}

\author{%
  Jue~Hou\textsuperscript{1,2,}\thanks{Equal Contributions.}, 
  Chengyi~Yang\textsuperscript{1,2,}\footnotemark[1], 
  Renxu~Bao\textsuperscript{1,2}, 
  Zhishang~Xiang\textsuperscript{1,2}, \\
  \textbf{Junnan~Zhu\textsuperscript{4}}, 
\textbf{Delai~Qiu\textsuperscript{3}}, 
\textbf{Yining~Wang\textsuperscript{3}}, 
\textbf{Shengping~Liu\textsuperscript{3}}, 
  \textbf{Jinsong~Su\textsuperscript{1,2,}}\thanks{Corresponding Author.}  \\
  \textsuperscript{1}Xiamen University \quad  \textsuperscript{2}Key Lab of Digital Protection and Intelligent Processing of\\ Intangible Cultural Heritage of Fujian-Taiwan (XMU), Ministry of Culture and Tourism \\
  \textsuperscript{3}Unisound\quad
  \textsuperscript{4}Institute of Automation, Chinese Academy of Sciences 
  \\
  \texttt{\{houjue,yangchengyi\}@stu.xmu.edu.cn}  \quad\texttt{jssu@xmu.edu.cn}
}

\begin{document}
\maketitle

\begin{abstract}
Equipping large language model agents with reusable skills improves
exploration and generalization in agentic reinforcement learning. However,
existing skill-learning methods typically distill trajectories into isolated
skills via costly language-model generation and update the bank only by
adding or removing whole skills, leaving cross-trajectory experience
underutilized and local flaws hard to revise. We propose \textbf{G-Skill}, a \textbf{graph-grounded skill evolution
framework} that represents accumulated experience as persistent task-specific
state-transition graphs rather than standalone skills. Through
\emph{Graph-Grounded Skill Construction}, G-Skill exploits shared states across
rollouts to derive dual-granularity \emph{Task Skills} and \emph{Step Skills}
from graph structure without additional language-model generation. Through
\emph{Skill Evolution}, new rollouts expand each graph so task-level procedures
and state-level action experience are continually refined in place and retrieved
to guide subsequent policy learning, letting skills and the policy co-evolve. Experiments on
ALFWorld and WebShop across two model families show that G-Skill outperforms
existing skill-based baselines and improves diverse reinforcement learning
methods, while ablations and analyses support its cross-trajectory path
recombination and fine-grained local refinement. Code is available at
\url{https://github.com/XMUDeepLIT/G-Skill}.
\end{abstract}

\section{Introduction}\label{sec:intro}

\begin{figure}[!t]
\centering
\includegraphics[width=.9\columnwidth]{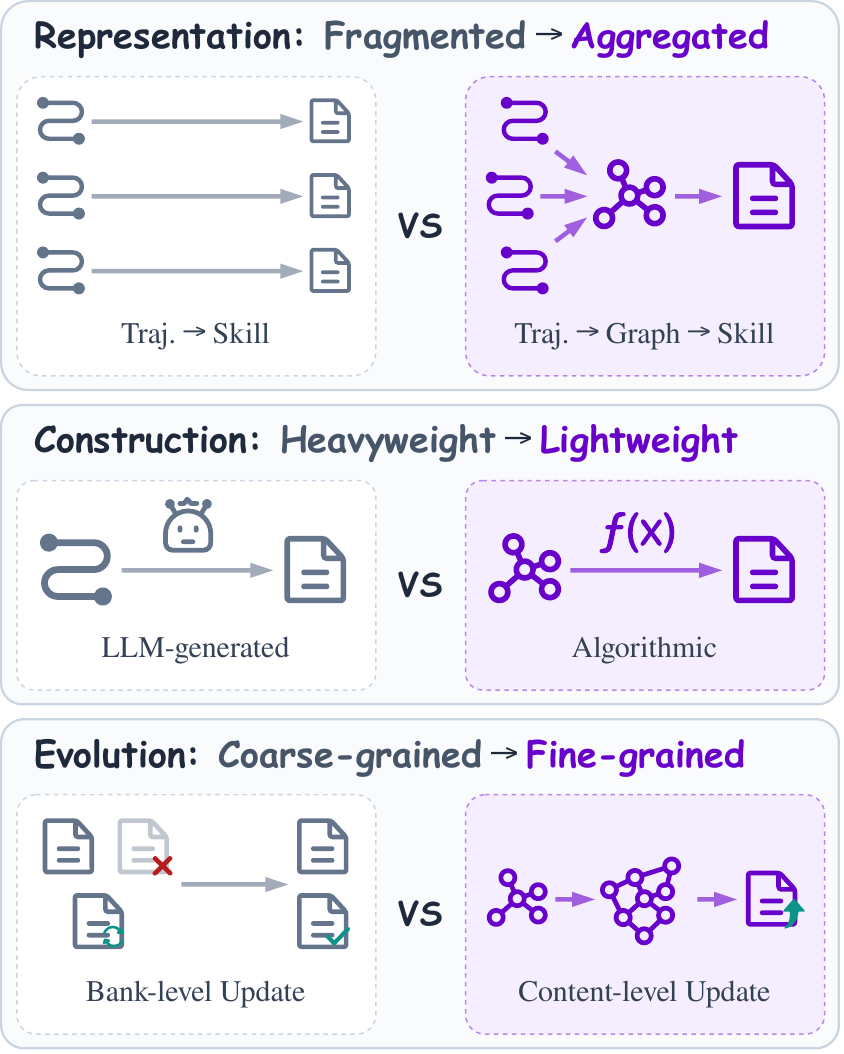}
\caption{\textbf{Comparison of skill learning.} Many existing methods store trajectories as isolated skills, generate them with a language model, and add or remove whole skills. In contrast, G-Skill merges trajectories into one graph, applies an algorithm, and revises skill content.}
\label{fig:motivation}
\end{figure}

Large language model (LLM) agents act iteratively on environment
feedback~\citep{yao2023react,shinn2023reflexion}, and reinforcement learning
(RL) further improves their policies via
interaction~\citep{wang2025ragen}. Yet learning
long-horizon procedures purely by trial and error is sample-inefficient and
repeatedly rediscovers common patterns. Reusable skills distilled from past
interactions address this bottleneck by transferring proven workflows and
local experience to new tasks, accelerating exploration and
generalization~\citep{wang2024voyager,wang2025workflow,xia2026skillrl,tu2026d2skill,shi2026skill1}.
This raises a key question: how can agents exploit complementary rollouts to
construct and continually refine reusable skills?

As illustrated in Figure~\ref{fig:motivation}, existing skill-learning methods
typically distill trajectories into a skill bank using language models and
update it at the bank level, facing three limitations in representation,
construction, and evolution.
(i)~\textbf{Fragmented representation}: extracting skills from single rollouts
or trajectory pairs discards links across
attempts~\citep{shi2026skill1,tu2026d2skill}, hiding where routes meet and
which shared-state action shortens the solution.
(ii)~\textbf{Heavyweight skill construction}: using a
teacher~\citep{xia2026skillrl} or policy~\citep{shi2026skill1} to generate
skills ties broader coverage to repeated LLM generation calls.
(iii)~\textbf{Coarse-grained evolution}: adding or removing whole
skills~\citep{tu2026d2skill,shi2026skill1} makes it hard to revise one flawed
recommendation while preserving steps that remain effective.

These limitations motivate an aggregated representation across shared states,
lightweight skill construction without LLM generation, and fine-grained
content evolution. Inspired by GraphGPO~\citep{cheng2026graphgpo}, which
merges rollouts within a batch into a state-transition graph for credit
assignment, and earlier graph-based navigation
agents~\citep{savinov2018sptm,eysenbach2019sorb}, we observe that persisting a
task-specific state-transition graph across rounds meets all three needs. Paths combine segments across attempts into shorter known routes, and branches at shared states expose candidate actions with goal distances. Incremental graph updates then refine both procedures and local experience.

Following this idea, we propose \textbf{G-Skill}, a \textbf{graph-grounded
skill evolution framework} that derives and refines skills directly from
persistent task graphs rather than storing standalone skills.
Reusing GraphGPO's aggregation across persistent task graphs, G-Skill performs
\emph{Graph-Grounded Skill Construction} via a single reverse breadth-first
search (BFS) without LLM generation. From the resulting goal distances,
\emph{Task Skills} capture shortest known paths as task-level procedures, and
\emph{Step Skills} score candidate actions by distance reduction as state-level
action experience. Through \emph{Skill Evolution}, new rollouts expand each
graph to shorten Task Skills and revise Step Skill rankings in place. Retrieved
skills then guide rollouts on similar tasks so skills and the policy co-evolve.

We evaluate G-Skill on ALFWorld and WebShop across Qwen2.5-3B-Instruct and
Qwen3-4B-Instruct-2507, achieving the highest observed mean success in every
RL group. With GRPO, WebShop success rises from 62.8\% to 89.4\% on Qwen2.5-3B
and ALFWorld success from 38.7\% to 77.6\% on Qwen3-4B, while pairing G-Skill
with GraphGPO yields further gains. Ablations support dual-level skills,
continual evolution, and relevance retrieval, and analyses contrast
long-horizon path recombination on ALFWorld with local Step Skill refinement
on short-horizon WebShop.

\section{Preliminary}\label{sec:preliminary}

\paragraph{Agentic Task Formulation.}
We consider an LLM agent completing a task instance $x\sim\mathcal{D}$ from
task distribution $\mathcal{D}$, with textual description $I$, through
interaction. At step $t\in\{1,\ldots,T\}$ of a $T$-step episode, the agent
observes state $s_t$ and takes action $a_t\sim\pi_\theta(\cdot\mid h_t)$,
where $\pi_\theta$ is a $\theta$-parameterized policy and $h_t$ is the history
containing $I$, $s_t$, and recent turns. An episode produces a trajectory
$\tau=(s_1,a_1,\ldots,s_T,a_T)$ with final task reward $r(\tau)$, maximizing
expected return
$J(\theta)=\mathbb{E}_{x\sim\mathcal{D},\,\tau\sim\pi_\theta(\cdot\mid x)}[r(\tau)]$.

\paragraph{Skills for LLM Agents.}
To reuse knowledge from past interactions, a skill-augmented agent maintains
a skill bank $\mathcal{B}=\{z_1,z_2,\ldots\}$, where each skill
$z\in\mathcal{B}$ encodes reusable interaction experience, from concrete task
procedures and state-level action patterns to abstract strategies, for future
decisions. At step $t$, a retrieval operator $\mathcal{R}$ selects relevant
skills $\mathcal{Z}_t=\mathcal{R}(h_t,\mathcal{B})\subseteq\mathcal{B}$, and
the policy samples $a_t\sim\pi_\theta(\cdot\mid h_t,\mathcal{Z}_t)$ under the
same reward objective. We study how an agent constructs $\mathcal{B}$ from
accumulated rollouts and continually refines its skills during RL training.

\section{Method}\label{sec:method}

\begin{figure*}[!t]
\centering
\includegraphics[width=\textwidth]{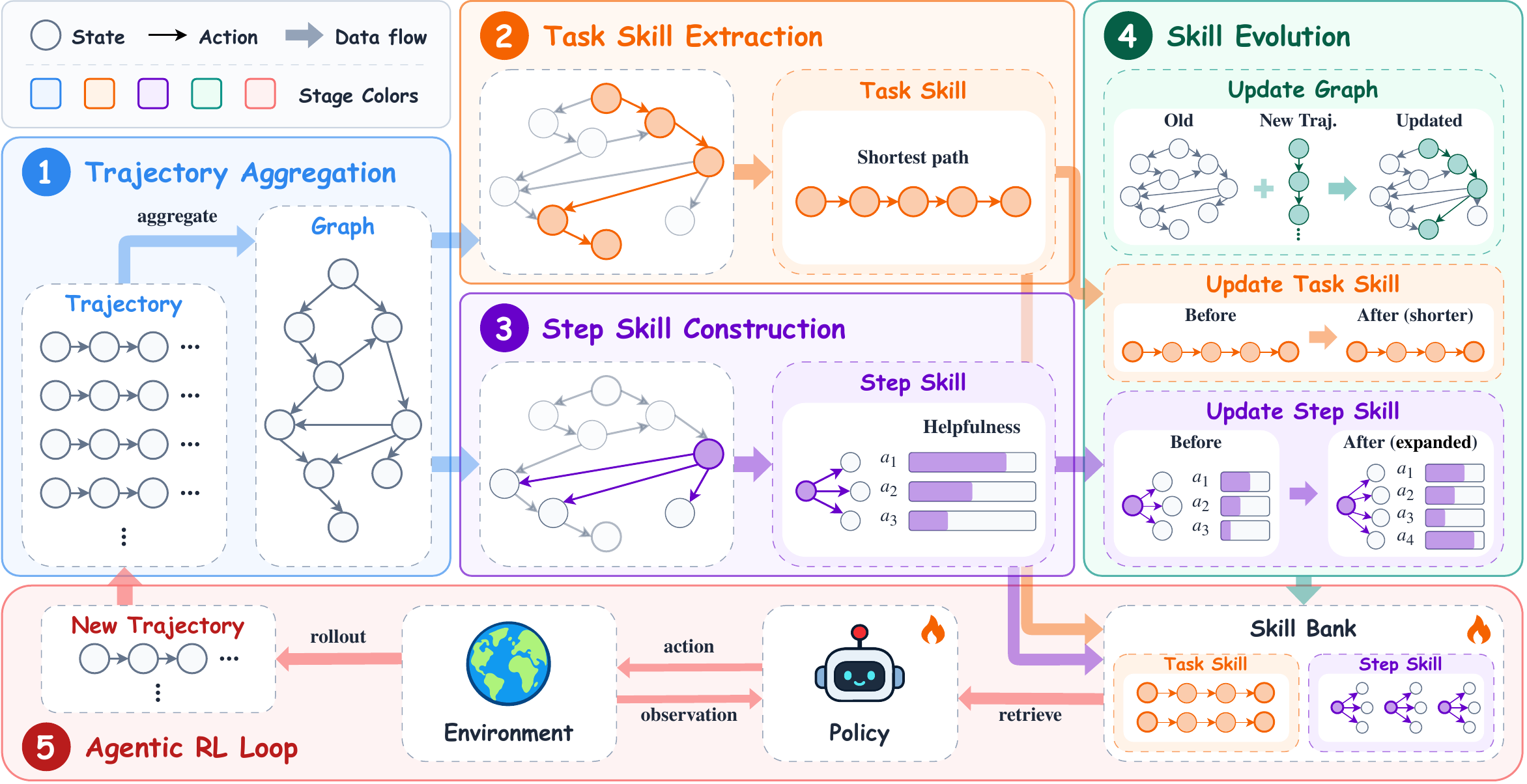}
\caption{
\textbf{Overview of G-Skill.}
\textcolor[HTML]{2E87EE}{\textbf{(1) Trajectory Aggregation}}: Multiple interaction trajectories are aggregated into a unified state-transition graph.
\textcolor[HTML]{F56301}{\textbf{(2) Task Skill Extraction}}: Task Skills are extracted from shortest known graph paths via a shared reverse BFS.
\textcolor[HTML]{6801CA}{\textbf{(3) Step Skill Construction}}: Step Skills are constructed from the same goal distances via candidate action helpfulness.
\textcolor[HTML]{065F46}{\textbf{(4) Skill Evolution}}: New trajectories update the graph and refine both types of skills.
\textcolor[HTML]{B91C1C}{\textbf{(5) Agentic RL Loop}}: Retrieved skills guide rollouts while environment feedback updates the policy.
}
\label{fig:main}
\end{figure*}

In this section, we propose G-Skill, a graph-grounded skill evolution
framework. Rather than distilling isolated skills with LLMs, it maintains
persistent task graphs and derives skills directly from their structure. As
shown in Figure~\ref{fig:main}, \emph{Trajectory Aggregation} merges task
rollouts into each graph (Section~\ref{sec:traj-agg}). \emph{Graph-Grounded
Skill Construction} derives two skill types via reverse BFS: \emph{Task Skill
Extraction} forms Task Skills from shortest known paths, and \emph{Step Skill
Construction} builds Step Skills by scoring actions via goal-distance
reduction (Section~\ref{sec:skill-const}). Next, \emph{Skill Evolution}
updates both skills in place as new rollouts expand each graph
(Section~\ref{sec:skill-evol}). In the \emph{Agentic RL Loop}, retrieved
skills guide rollouts that update the policy (Section~\ref{sec:rl-loop}).
Repeating this loop lets policy and skills co-evolve from the same experience.

\subsection{Trajectory Aggregation}\label{sec:traj-agg}
For each task instance $x$, we aggregate collected rollouts into a directed
graph $\mathcal{G}=(\mathcal{S},\mathcal{E})$, following
GraphGPO~\citep{cheng2026graphgpo}. The node set $\mathcal{S}$ holds all
visited states and terminal goal node $s^*$, with all rollouts sharing initial
state $s_1$. Each edge $(s,a,s')\in\mathcal{E}$ records an observed transition with
environment-normalized action $a$, and successful rollouts end with
$(s_T,a_T,s^*)$. Nonterminal transitions from both successful and failed
rollouts are retained. State merging occurs strictly within the same
task instance under an environment-specific state representation, so path
executability rests on retaining transition-relevant state details, as
detailed in Appendices~\ref{app:state-rep} and~\ref{app:graph-skill-impl}.

Each task graph persists across rounds: rollouts for $x$ in its $n$-th round
form an increment $\Delta\mathcal{G}^{(n)}$, updating graph $\mathcal{G}^{(n)}$
from empty initial graph $\mathcal{G}^{(0)}=\varnothing$:
\begin{equation}
 \mathcal{G}^{(n)}=
 \mathcal{G}^{(n-1)}\cup\Delta\mathcal{G}^{(n)},
 \quad \mathcal{G}^{(0)}=\varnothing.
 \label{eq:graph-update}
\end{equation}

\subsection{Graph-Grounded Skill Construction}\label{sec:skill-const}
To extract reusable experience, we quantify each state's shortest path to
success. Let $\mathcal{P}(s)$ be the directed paths from $s$ to $s^*$ with
transition count $|p|$, and define goal distance $d(s)$ as
\begin{equation}
 d(s)=
 \begin{cases}
 \min\limits_{p\in\mathcal{P}(s)}|p|,
   & \mathcal{P}(s)\neq\varnothing,\\
 +\infty, & \mathcal{P}(s)=\varnothing,
 \end{cases}
 \label{eq:distance}
\end{equation}
We compute $d(s)$ via a single reverse BFS from $s^*$ with $d(s^*)=0$, so
finite $d(s_1)$ is the shortest path length to $s^*$. Subtracting successor
from current distance measures progress toward $s^*$.

\paragraph{Task Skill Extraction.}
If $d(s_1)<+\infty$, we select a shortest path $p\in\mathcal{P}(s_1)$ with
$|p|=d(s_1)$:
\begin{equation}
 p=(s_1,a_1,s_2,a_2,\ldots,s_{|p|},a_{|p|},s^*).
\end{equation}
The Task Skill pairs the task description with this shortest known graph path,
$z_{\mathrm{task}}=(I,p)$, where each edge records an action and its outcome.
At shared states, $p$ can combine shorter continuations from different
rollouts, avoiding detours.

\paragraph{Step Skill Construction.}
A Step Skill provides action examples from a similar state. An observed action
$a$ at state $s$ enters the eligible set $\mathcal{A}(s)$ if some observed
successor state $s'$ can reach $s^*$:
\begin{equation}
 \mathcal{A}(s)=
 \{a\mid(s,a,s')\in\mathcal{E},\ d(s')<+\infty\}.
\end{equation}
For each eligible action $a\in\mathcal{A}(s)$ at state $s$, we define its
helpfulness $H(s,a)$ as the maximum goal-distance reduction across its
finite-distance transitions:
\begin{equation}
 H(s,a)=\max_{\substack{(s,a,s')\in\mathcal{E}\\ d(s')<+\infty}}\bigl[d(s)-d(s')\bigr].
 \label{eq:helpfulness}
\end{equation}
With unit edge costs, $H(s,a)\leq 1$. Actions starting a shortest known path
score $1$, a score of $0$ leaves remaining distance unchanged, and $-k$ moves
$k$ steps farther from $s^*$ in the reference graph. Thus the maximum gives an optimistic estimate of observed progress in the
reference graph rather than expected task utility, with action-normalization
rules detailed in Appendix~\ref{app:graph-skill-impl}.

For any non-goal state $s$ with $d(s)<+\infty$, at least one action in
$\mathcal{A}(s)$ lies on a shortest path and has helpfulness $1$. We rank the
eligible actions by helpfulness and retain up to $K$ action--score pairs in
the ranked list $C(s)$:
\begin{equation}
 C(s)=\operatorname{TopK}
 \{(a,H(s,a))\mid a\in\mathcal{A}(s)\}.
\end{equation}
Pairing $I$, $s$, and $C(s)$ forms Step Skill $z_{\mathrm{step}}(s)=(I,s,C(s))$,
where tied scores mean equal progress and higher scores mean shorter paths.

\begin{figure}[!t]
    \centering
    \includegraphics[width=\columnwidth]{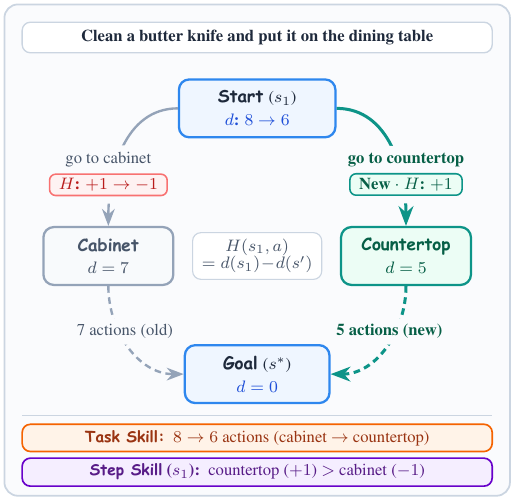}
    \caption{\textbf{An ALFWorld shortcut updates action helpfulness.}
    Nodes show shortest distance $d$ to goal $s^*$, and badges show helpfulness $H(s_1,a)=d(s_1)-d(s')$. Dashed arrows denote action sequences.}
    \label{fig:skill-evolution-case}
\end{figure}

\subsection{Skill Evolution}\label{sec:skill-evol}
When new rollouts enter a task graph via Equation~\ref{eq:graph-update}, we
recompute goal distances and mark updated quantities with superscript $n$.
Because graph updates only add nodes and edges, goal distances are
monotonically non-increasing, $d^{(n)}(s)\leq d^{(n-1)}(s)$. By contrast, action helpfulness can change non-monotonically because source and
successor distances may decrease unevenly. Existing Task Skills never lengthen,
whereas prior optimal actions can drop in rank as shortcuts appear.

\paragraph{Task Skill Evolution.}
A decrease $d^{(n)}(s_1)<d^{(n-1)}(s_1)$ creates the first Task Skill once
$s^*$ becomes reachable or replaces it with a shorter path, while ties follow
the tie-breaking rule.

\paragraph{Step Skill Evolution.}
A non-goal state gains a Step Skill once a path to $s^*$ is established. At an
existing skill state, actions become eligible as successors reach $s^*$ and
receive updated scores as distances change, re-ranking $C^{(n)}(s)$.

Figure~\ref{fig:skill-evolution-case} illustrates two successive ALFWorld
rounds: the Task Skill switches from the earlier eight-action cabinet route to
a six-action countertop route. At the initial state, going to the countertop
scores $+1$, whereas going to the cabinet drops from $+1$ to $-1$ since $s^*$
is seven actions away.

\subsection{Agentic RL Loop}\label{sec:rl-loop}
\paragraph{Skill Retrieval and Injection.}
Task Skills and Step Skills are retrieved separately using normalized BM25
similarity. A Task Skill is ranked by similarity between the current task description $I$
and each candidate's reference task description. A Step Skill averages
task-description similarity and state similarity between $s_t$ and the
candidate's reference state. We restrict candidates to the same task type, exclude identical task
descriptions to prevent exact-task leakage, and select up to one Task Skill
and two Step Skills to form $\mathcal{Z}_t$. Fixed templates prepend the Task
Skill subgoal order and Step Skill action examples before the current task,
with full retrieval formulas and prompt templates provided in
Appendices~\ref{app:retrieval-impl} and~\ref{app:prompt-templates}.

\paragraph{Policy Optimization.}
Conditioned on $\mathcal{Z}_t$, the policy samples
$a_t{\sim}\pi_\theta(\cdot{\mid}h_t,\mathcal{Z}_t)$. Each round holds the skill bank fixed while collecting $N$ rollouts per task.
These rollouts merge into each task graph to update both skills and also update
$\pi_\theta$ with the base RL objective, as in Algorithm~\ref{alg:gskill}.

\section{Experiments}\label{sec:exp}

\begin{table*}[!t]
\centering
\resizebox{\textwidth}{!}{%
\begin{tabular}{lccccccc|cc}
\toprule
\multirow{2}{*}{\textbf{Method}}
& \multicolumn{7}{c|}{\textbf{ALFWorld}}
& \multicolumn{2}{c}{\textbf{WebShop}} \\
\cmidrule(lr){2-8}\cmidrule(lr){9-10}
& Pick & Clean & Cool & Look & Heat & Pick2 & All & Score & Succ. \\
\midrule
\rowcolor{gray!15}
\multicolumn{10}{c}{\textbf{Qwen2.5-3B-Instruct}} \\
\cmidrule{1-10}
\rowcolor{gray!6}
\multicolumn{10}{l}{\textit{w/o RL Training}} \\
\quad Base Model
& 55.4\std{6.6} & 11.1\std{3.7} & 6.4\std{2.2}
& 32.3\std{6.4} & 8.8\std{7.1} & 7.5\std{1.9}
& 22.4\std{1.6} & 6.1\std{1.3} & 2.5\std{1.8} \\
\addlinespace[2pt]
\rowcolor{gray!6}
\multicolumn{10}{l}{\textit{GRPO-based Methods}} \\
\quad GRPO~\citep{shao2024deepseekmath}
& 93.7\std{3.7} & 93.3\std{3.1} & 66.4\std{9.6}
& 64.6\std{6.9} & 81.3\std{6.3} & 48.3\std{7.0}
& 76.9\std{3.0} & 74.9\std{3.3} & 62.8\std{3.4} \\
\quad SkillRL~\citep{xia2026skillrl}
& 77.1\std{5.0} & 76.3\std{10.7} & 68.8\std{3.4}
& 61.5\std{7.7} & 47.5\std{5.6} & 50.8\std{5.4}
& 66.1\std{1.8} & 70.4\std{2.3} & 55.3\std{2.1} \\
\quad D2Skill~\citep{tu2026d2skill}
& 89.1\std{3.7} & 77.8\std{9.4} & 60.8\std{8.2}
& 67.7\std{6.4} & 75.0\std{4.4} & 55.0\std{10.4}
& 72.4\std{1.6} & 71.9\std{2.4} & 54.4\std{4.7} \\
\quad Skill1~\citep{shi2026skill1}
& \textbf{96.6\std{2.4}} & \textbf{94.1\std{4.2}} & 58.4\std{5.4}
& 66.2\std{6.9} & \textbf{93.8\std{4.4}} & 77.5\std{4.8}
& 82.9\std{0.7} & 50.4\std{2.1} & 37.8\std{2.3} \\
\quad \textbf{G-Skill} w/ GRPO
& 84.0\std{2.6} & 86.7\std{3.3} & \textbf{81.6\std{2.2}}
& \textbf{76.9\std{5.4}} & 85.0\std{3.4} & \textbf{83.3\std{4.2}}
& \textbf{83.4\std{1.2}} & \textbf{92.6\std{2.0}} & \textbf{89.4\std{2.0}} \\
\addlinespace[2pt]
\rowcolor{gray!6}
\multicolumn{10}{l}{\textit{GraphGPO-based Methods}} \\
\quad GraphGPO~\citep{cheng2026graphgpo}
& \textbf{93.7\std{1.3}} & \textbf{94.8\std{2.0}} & 83.2\std{1.8}
& 69.2\std{7.7} & 77.5\std{5.6} & \textbf{83.3\std{4.2}}
& 86.1\std{2.1} & 83.3\std{1.9} & 73.8\std{2.5} \\
\quad \textbf{G-Skill} w/ GraphGPO
& 88.6\std{2.0} & 92.6\std{2.6} & \textbf{90.4\std{2.2}}
& \textbf{86.2\std{3.4}} & \textbf{91.3\std{3.4}} & 77.5\std{2.3} & \textbf{87.9\std{0.9}}
& \textbf{97.0\std{1.1}} & \textbf{92.8\std{1.4}} \\
\midrule
\rowcolor{gray!15}
\multicolumn{10}{c}{\textbf{Qwen3-4B-Instruct-2507}} \\
\cmidrule{1-10}
\rowcolor{gray!6}
\multicolumn{10}{l}{\textit{w/o RL Training}} \\
\quad Base Model
& 58.9\std{3.3} & 13.3\std{2.0} & 6.4\std{4.6}
& 23.1\std{7.7} & 32.5\std{6.9} & 10.0\std{2.3}
& 26.0\std{1.0} & 16.4\std{5.0} & 3.4\std{1.7} \\
\addlinespace[2pt]
\rowcolor{gray!6}
\multicolumn{10}{l}{\textit{GRPO-based Methods}} \\
\quad GRPO~\citep{shao2024deepseekmath}
& 80.0\std{4.5} & 23.7\std{4.2} & 15.2\std{3.4}
& 21.5\std{10.0} & 43.8\std{4.4} & 25.8\std{5.4}
& 38.7\std{2.3} & 60.5\std{15.9} & 48.4\std{10.5} \\
\quad SkillRL~\citep{xia2026skillrl}
& 81.7\std{3.3} & 60.0\std{3.1} & 59.2\std{7.7}
& \textbf{76.9\std{0.0}} & 36.3\std{2.8} & 43.3\std{11.3}
& 61.3\std{4.2} & 65.3\std{10.8} & 53.4\std{8.5} \\
\quad D2Skill~\citep{tu2026d2skill}
& 88.6\std{2.0} & 75.6\std{5.0} & 55.2\std{5.2}
& 56.9\std{4.2} & 71.3\std{9.5} & 25.0\std{6.6}
& 64.3\std{1.2} & 59.8\std{5.5} & 45.3\std{4.6} \\
\quad Skill1~\citep{shi2026skill1}
& 64.6\std{3.3} & 41.5\std{5.5} & 38.4\std{5.4}
& 56.9\std{4.2} & 53.8\std{7.1} & 40.8\std{5.4}
& 49.4\std{1.9} & 23.3\std{2.7} & 10.6\std{2.6} \\
\quad \textbf{G-Skill} w/ GRPO
& \textbf{89.1\std{3.7}} & \textbf{83.7\std{2.0}} & \textbf{76.0\std{0.0}}
& 49.2\std{4.2} & \textbf{73.8\std{5.2}} & \textbf{73.3\std{5.6}}
& \textbf{77.6\std{2.3}} & \textbf{86.7\std{1.5}} & \textbf{74.7\std{2.8}} \\
\addlinespace[2pt]
\rowcolor{gray!6}
\multicolumn{10}{l}{\textit{GraphGPO-based Methods}} \\
\quad GraphGPO~\citep{cheng2026graphgpo}
& \textbf{97.1\std{0.0}} & 93.3\std{3.1} & \textbf{78.4\std{2.2}}
& 81.5\std{11.7} & 83.8\std{5.6} & 86.7\std{4.6}
& 88.3\std{0.8} & 65.0\std{2.4} & 55.3\std{1.4} \\
\quad \textbf{G-Skill} w/ GraphGPO
& 96.6\std{2.4} & \textbf{94.8\std{2.0}} & \textbf{78.4\std{2.2}}
& \textbf{100.0\std{0.0}} & \textbf{86.3\std{5.2}} & \textbf{95.0\std{1.9}}
& \textbf{91.9\std{1.1}} & \textbf{91.6\std{1.6}} & \textbf{83.4\std{0.9}} \\
\bottomrule
\end{tabular}%
}
\caption{\textbf{Main test performance on ALFWorld and WebShop.} For ALFWorld, we report
success rate (\%) per subtask and overall, and for WebShop, average task score
scaled to 0--100 and success rate (\%). Results average five evaluation
sampling seeds per checkpoint, with standard deviations after $\pm$. Best per
RL group is in \textbf{bold}.}
\label{tab:main_results}
\end{table*}

We organize our experiments around four research questions:
\textbf{RQ1:} How effective is G-Skill compared with existing RL and skill-based methods?
\textbf{RQ2:} Which components of G-Skill contribute to its performance?
\textbf{RQ3:} How broadly does G-Skill cover source tasks, and how similar are its available references to evaluation tasks?
\textbf{RQ4:} How do lightweight construction, graph aggregation, and fine-grained evolution operate in practice?

\subsection{Experimental Setup}

\paragraph{Environments.}
We evaluate G-Skill on ALFWorld~\citep{shridhar2021alfworld} and
WebShop~\citep{yao2022webshop}. In ALFWorld, agents solve
ALFRED~\citep{shridhar2020alfred} tasks via
TextWorld~\citep{cote2018textworld} across 140 \texttt{valid\_seen} games. In
WebShop, agents search and buy products across 64 benchmark slots, with full
details in Appendix~\ref{app:environments}.

\paragraph{Baselines.}
We compare G-Skill with skill-free and skill-augmented RL methods. Base Model
is the initial checkpoint without RL. For skill-free RL, we include
GRPO~\citep{shao2024deepseekmath} and GraphGPO~\citep{cheng2026graphgpo}. For
skill-augmented RL, we include SkillRL~\citep{xia2026skillrl},
D2Skill~\citep{tu2026d2skill}, and Skill1~\citep{shi2026skill1}, detailed in
Appendix~\ref{app:baselines}.

\paragraph{Implementation Details.}
We conduct experiments with Qwen2.5-3B-Instruct~\citep{qwen2025qwen25}
and Qwen3-4B-Instruct-2507~\citep{yang2025qwen3}.
We train for 140 steps with a learning rate of $1\times10^{-6}$,
16 tasks per batch, and eight rollouts per task.
For WebShop experiments with Qwen3-4B-Instruct-2507, all methods start from
the same checkpoint fine-tuned on the open-source data released by
SkillRL~\citep{xia2026skillrl}. During evaluation, all skill-augmented methods use their skill banks frozen after training without test-time updates. G-Skill builds its bank exclusively from
training rollouts, while baselines keep their standard protocols, detailed in
Appendix~\ref{app:training-config}.

\begin{table*}[!t]
\centering
\begin{tabular}{lccccc|cc}
\toprule
\multirow{2}{*}{\textbf{Method}}
& \multicolumn{5}{c|}{\textbf{G-Skill Components}}
& \multicolumn{2}{c}{\textbf{WebShop}} \\
\cmidrule(lr){2-6}\cmidrule(lr){7-8}
& Graph Agg. & Task & Step & Evolution & Retrieval & Score & Succ. \\
\midrule
GRPO
& -- & -- & -- & -- & --
& 74.9\std{3.3} & 62.8\std{3.4} \\
Single-Trajectory Skill
& -- & $\checkmark$ & -- & -- & BM25
& 90.1\std{0.8} & 85.3\std{0.9} \\
\midrule
\textbf{G-Skill}
& $\checkmark$ & $\checkmark$ & $\checkmark$ & $\checkmark$ & BM25
& \textbf{92.6\std{2.0}} & \textbf{89.4\std{2.0}} \\
\quad w/o Task Skill
& $\checkmark$ & -- & $\checkmark$ & $\checkmark$ & BM25
& 89.6\std{3.9} & 84.1\std{2.8} \\
\quad w/o Step Skill
& $\checkmark$ & $\checkmark$ & -- & $\checkmark$ & BM25
& 91.4\std{1.3} & 84.1\std{2.8} \\
\quad w/o Graph Evolution
& $\checkmark$ & $\checkmark$ & $\checkmark$ & -- & BM25
& 89.3\std{1.4} & 82.5\std{1.3} \\
\quad w/ Random Retrieval
& $\checkmark$ & $\checkmark$ & $\checkmark$ & $\checkmark$ & Random
& 75.1\std{3.6} & 69.1\std{3.7} \\
\bottomrule
\end{tabular}
\caption{\textbf{Ablation study of G-Skill on WebShop with Qwen2.5-3B and GRPO.}
Checkmarks ($\checkmark$) mark enabled components and retrieval rule. Score (0--100) and Succ.\ (\%) report five-seed means $\pm$ SD, with best in \textbf{bold}.}
\label{tab:skill_ablation}
\end{table*}

\subsection{Main Results (RQ1)}

Table~\ref{tab:main_results} reports the results on ALFWorld and WebShop.
Each mean averages five evaluation runs of a single checkpoint, where G-Skill
achieves the highest observed mean overall success in each RL group across
both benchmarks and base models.

\paragraph{G-Skill outperforms GRPO and LLM-generated skill baselines.}
On WebShop, G-Skill leads the GRPO group on both task score and success,
lifting Qwen2.5-3B success from 62.8\% with GRPO to 89.4\%. On ALFWorld, the gain over GRPO is larger on Qwen3-4B. On Qwen2.5-3B, G-Skill
and Skill1 reach similar overall success yet lead different families, with
G-Skill ahead on Cool and Pick2 and Skill1 on Pick and Clean. Notably, existing
LLM-generated skill baselines occasionally underperform vanilla GRPO under the
shared training budget. Skill1 suffers structured self-distillation and rerank
formatting collapse on WebShop, whereas teacher-driven banks face heavy
eviction churn in D2Skill or early saturation in SkillRL, as diagnosed in
Appendix~\ref{app:baseline-analysis}.

\begin{figure*}[!t]
    \centering
    \includegraphics[width=.94\textwidth]{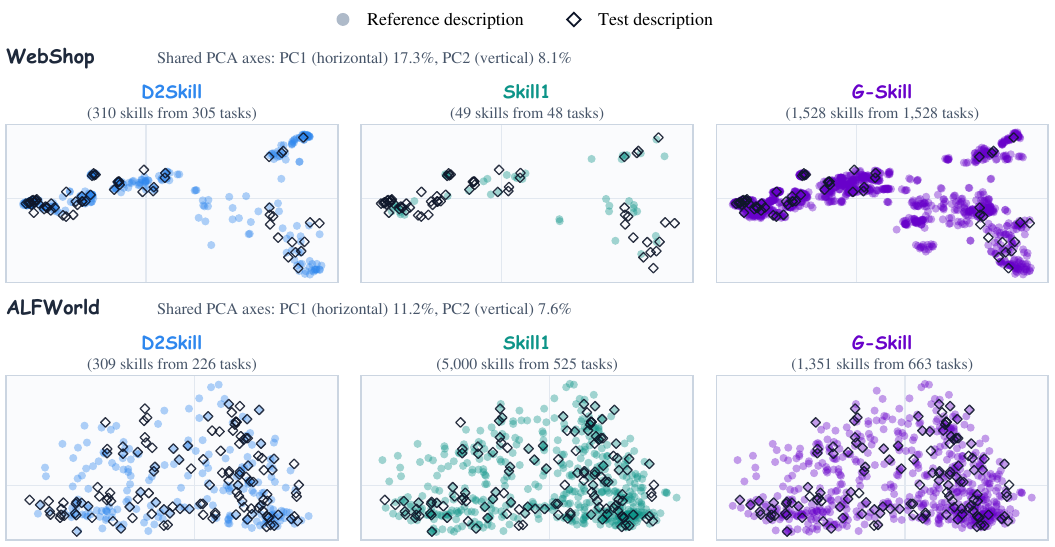}
    \caption{\textbf{Task-level coverage of Qwen2.5-3B evaluation banks.}
    Dots show distinct source task descriptions and hollow diamonds show test tasks on shared Qwen3-Embedding-0.6B PCA axes as a qualitative overview. SkillRL is omitted because its abstract principles lack single source tasks. Subtitles list total Task Skills and distinct source descriptions, showing that G-Skill retains the most distinct source descriptions in both environments.}
    \label{fig:reference-task-distribution}
\end{figure*}

\paragraph{G-Skill complements graph-based credit assignment in GraphGPO.}
Even though GraphGPO already uses per-batch graphs for credit assignment,
adding skills from persistent task graphs raises its mean success on
both benchmarks and models, with larger gains and higher task scores on
WebShop. Thus, retrieved skills complement batch-level credit assignment.

\subsection{Ablation Study (RQ2)}

We conduct ablations on WebShop using Qwen2.5-3B-Instruct with GRPO under the
same budget and protocol. Table~\ref{tab:skill_ablation} compares G-Skill with
variants of skill construction, evolution, and retrieval, with full variant
configurations detailed in Appendix~\ref{app:ablation-configs}.

\paragraph{Combining Task and Step Skills outperforms single-level skills.}
Single-Trajectory Skill presents one successful rollout, whereas G-Skill w/o
Step Skill extracts a shortest graph path under the same format and retrieval
rule. On short-horizon WebShop, where cross-trajectory shortcuts are rare at
0.4\% in Table~\ref{tab:skill-evolution}, graph-derived Task Skills alone do
not exceed Single-Trajectory Skill, reaching 84.1\% versus 85.3\% success with
a slightly higher task score. Combining task-level procedures with state-level
Step Skills yields the highest observed success at 89.4\%, versus 84.1\% for
either skill alone under the same graph evolution mechanism. Step Skills also
expose alternative actions when multiple eligible patterns exist.

\paragraph{Continual graph evolution is essential for sustained gains.}
Freezing each task graph after its first rollout group retains a comparable
number of Task Skills but blocks subsequent goal links, candidate actions, and
score revisions, lowering WebShop success from 89.4\% to 82.5\%.

\paragraph{Relevance-based retrieval drives most of the performance gain.}
Sampling randomly among eligible skills causes the largest drop from full
G-Skill, pulling task score close to GRPO and indicating the importance of
matching retrieved references to the current task and state.

\subsection{Skill Bank Coverage and Reference Similarity (RQ3)}
\label{sec:skill-distribution}

\begin{figure}[!t]
    \centering
    \includegraphics[width=\columnwidth]{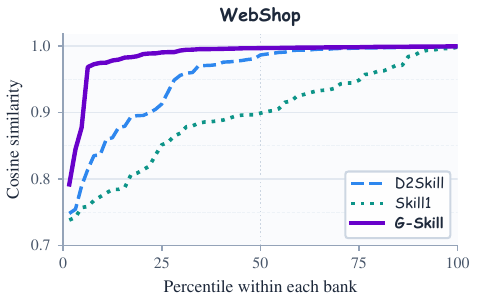}
    \caption{\textbf{Closest non-identical reference similarity on WebShop with Qwen2.5-3B.}
    For all 64 test task descriptions, we compute Qwen3-Embedding-0.6B cosine similarity against non-identical source tasks in each bank. Curves plot each task's highest score by percentile (x-axis), where higher curves show closer matches.}
    \label{fig:nearest-reference}
\end{figure}

\paragraph{G-Skill retains broader unique task coverage than baselines.}
Figure~\ref{fig:reference-task-distribution} compares distinct source task coverage across Qwen2.5-3B evaluation banks, omitting SkillRL because its abstract principles do not map to single training tasks.
On WebShop, G-Skill retains 1,528 Task Skills from 1,528 distinct descriptions, versus 305 for D2Skill and 48 for Skill1.
On ALFWorld, raw skill counts can mislead: Skill1 holds 5,000 Task Skills from 525 descriptions, whereas G-Skill spans 663 descriptions in 1,351 Task Skills plus far more Step Skills in Table~\ref{tab:construction-cost}.

\paragraph{G-Skill provides closer stored references for unseen test tasks.}
To measure reference coverage beyond BM25, we compute Qwen3-Embedding-0.6B cosine similarity between each test task and non-identical source tasks in each bank, taking the highest score as its nearest reference.
On WebShop in Figure~\ref{fig:nearest-reference}, G-Skill leads across all quantiles, especially on requests with the weakest matches. Table~\ref{tab:nearest-reference-webshop} compares each bank's closest task on a typical T-shirt request (nearest to all three banks' medians). G-Skill preserves size and price cap, whereas baselines miss all three attributes (Appendices~\ref{app:reference-map} and~\ref{app:alfworld-reference}).

\subsection{Construction Efficiency, Aggregation, and Evolution (RQ4)}
\label{sec:skill-benefits}

\paragraph{Graph-based construction removes LLM overhead with millisecond BFS.}
Table~\ref{tab:construction-cost} isolates skill construction from RL updates and retrieval.
SkillRL and D2Skill rely on external teacher LLM calls, with D2Skill evicting 502 ALFWorld and 950 WebShop skills during training, while Skill1 spends extra on-policy decoding turns.
G-Skill eliminates additional LLM generation calls for skill construction,
computes graph distances via a linear-time $O(|\mathcal{S}|+|\mathcal{E}|)$
reverse BFS, and materializes skills using fixed templates.
On CPU, reverse BFS takes 118.6\,ms over 1,674 ALFWorld graphs (0.071\,ms/graph) and 63.4\,ms over 2,083 WebShop graphs (0.030\,ms/graph), building larger banks without API latency or decoding collapse.

\begin{table}[!t]
\centering
\footnotesize
\begin{tabular*}{\columnwidth}{@{\extracolsep{\fill}}llll@{}}
\toprule
\textbf{Source} & \textbf{Color} & \textbf{Size} & \textbf{Price} \\
\midrule
\textbf{Test task} & \textbf{Pink} & \textbf{XXL} & \textbf{\textless\,\$40} \\
\midrule
D2Skill & Heather Grey & XS & \textless\,\$50 \\
Skill1  & Fish & XXL Big Tall & \textless\,\$50 \\
\textbf{G-Skill} & Baby Blue & \textbf{XXL} & \textbf{\textless\,\$40} \\
\bottomrule
\end{tabular*}
\caption{\textbf{Nearest stored reference tasks for a typical WebShop T-shirt test request.}
Chosen as the test request whose similarities across the three banks lie closest to each bank's median. Each row gives a bank's closest non-identical source task by cosine similarity, with matched attributes in \textbf{bold}.}
\label{tab:nearest-reference-webshop}
\end{table}

\begin{table*}[!t]
\centering
\footnotesize
\resizebox{\textwidth}{!}{%
\begin{tabular}{@{}llllrr@{}}
\toprule
\textbf{Method} & \textbf{Construction Mechanism} & \textbf{LLM Dependency} & \textbf{Main Construction Operation} & \textbf{ALFWorld Bank} & \textbf{WebShop Bank} \\
\midrule
SkillRL & Teacher LLM distillation & Teacher (24 / 9 rounds) & API latency & 127 principles & 93 principles \\
D2Skill & Teacher reflection + pruning & Teacher (502 / 950 evicted) & API latency & 309 T + 309 S & 310 T + 311 S \\
Skill1  & On-policy LLM distillation & Extra policy turns & Policy decoding & 5,000 T & 49 T \\
\midrule
\textbf{G-Skill} & \textbf{Reverse BFS + templates} & \textbf{None (0 calls)} & \textbf{$O(|\mathcal{S}|+|\mathcal{E}|)$ (0.071 / 0.030\,ms)} & \textbf{1,351 T + 44,982 S} & \textbf{1,528 T + 25,652 S} \\
\bottomrule
\end{tabular}%
}
\caption{\textbf{Skill construction comparison on Qwen2.5-3B during training.}
This table isolates construction from RL updates and retrieval. ``A / B'' pairs give ALFWorld / WebShop values (SkillRL rounds, D2Skill evictions, and G-Skill per-graph BFS time). ``T'' and ``S'' denote Task and Step Skills. Pure CPU BFS takes 118.6\,ms over 1,674 ALFWorld graphs and 63.4\,ms over 2,083 WebShop graphs, excluding maintenance and indexing.}
\label{tab:construction-cost}
\end{table*}

\begin{table}[!t]
\centering
\footnotesize
\renewcommand{\arraystretch}{0.92}
\resizebox{\columnwidth}{!}{%
\begin{tabular}{@{}l@{\hspace{6pt}}r@{\hspace{8pt}}r@{}}
\toprule
\textbf{Mechanism / Metric} & \textbf{ALFWorld} & \textbf{WebShop} \\
\midrule
\multicolumn{3}{@{}l}{\textit{Within-graph aggregation}} \\
\quad Shorter than any successful rollout & 410 / 1,351 (30.3\%) & 6 / 1,528 (0.4\%) \\
\quad Max actions saved & 8 & 2 \\
\midrule
\multicolumn{3}{@{}l}{\textit{Cross-round graph growth}} \\
\quad Task / Step Skills created & 1,351 / 44,982 & 1,528 / 25,652 \\
\quad Step Skills via new goal links & 2,233 & 1,308 \\
\midrule
\multicolumn{3}{@{}l}{\textit{Fine-grained content revision (revisited tasks)}} \\
\quad Task Skills shortened / eligible & 103 / 344 (29.9\%) & 1 / 100 (1.0\%) \\
\quad Step Skills revised / eligible & 2,877 / 11,262 (25.5\%) & 258 / 1,992 (13.0\%) \\
\quad \hspace{0.5em}w/ Task Skill text unchanged & 1,665 (57.9\%) & 126 (48.8\%) \\
\quad Candidate pools w/ added actions & 2,560 & 287 \\
\quad Candidate pools w/ changed scores & 1,532 & 62 \\
\bottomrule
\end{tabular}%
}
\caption{\textbf{Graph aggregation and evolution during training on Qwen2.5-3B with GRPO.}
Eligible skills belong to tasks resampled in a later round. ``w/ Task Skill text unchanged'' gives the share of revised Step Skills. Step Skill revisions count changed Top-3 lists, and candidate pools track full sets $\mathcal{A}(s)$.}
\label{tab:skill-evolution}
\end{table}

\paragraph{Persistent graphs drive recombination on ALFWorld and local edits on WebShop.}
Table~\ref{tab:skill-evolution} contrasts how persistent graphs operate across the two environments during Qwen2.5-3B GRPO training.
On long-horizon ALFWorld, within-graph aggregation produces paths shorter than every corresponding successful rollout on 30.3\% of Task Skills, saving up to 8 actions.
Across rounds, later rollouts connect 2,233 prior ALFWorld states and 1,308 WebShop states to the goal.
Among revisited ALFWorld tasks, G-Skill shortens 29.9\% of eligible Task Skills and revises 25.5\% of eligible Top-3 Step Skills.
By contrast, WebShop's 3--4 step episodes and rare task repeats leave little room for path shortening (1.0\%). Graph evolution therefore acts mainly through newly linked states and local Step Skill updates: 287 candidate pools gain actions, 62 change scores, and 13.0\% of eligible Top-3 Step Skills change. This local refinement aligns with the drop from 89.4\% to 82.5\% when freezing graphs in Table~\ref{tab:skill_ablation}.
Crucially, 57.9\% of revised Step Skills on ALFWorld and 48.8\% on WebShop refine local experience while the Task Skill text stays unchanged. Transfer failures and displayed action-comparison breakdowns are further analyzed in Appendices~\ref{app:skill-transfer-failures} and~\ref{app:skill-updates}.

\section{Related Work}\label{sec:related_work}

\paragraph{Learning reusable skills from experience.}
Early experience-reuse methods improve agents via failure
reflection~\citep{shinn2023reflexion} or cross-task
retrieval~\mbox{\citep{zhao2024expel}}. Later work abstracts interactions into
workflows, memory, and principles, including web
workflows~\citep{wang2025workflow}, procedural memory~\citep{fang2026memp},
self-distilled principles~\citep{wu2026evolver}, success--failure
distillation~\citep{ouyang2026reasoningbank}, and subtask-progress rewards with prior
lessons~\citep{zhang2026retroagent}. Closer to our setting, agentic RL maintains explicit skill banks.
SkillRL distills general and task-specific advice~\mbox{\citep{xia2026skillrl}},
D2Skill pairs task plans with step corrections and utility
pruning~\citep{tu2026d2skill}, and Skill1 trains one policy to select, use, and
generate skills~\citep{shi2026skill1}.
Unlike LLM-generated banks, G-Skill derives both skills from a
persistent task graph and updates procedures and action scores in place.

\paragraph{Credit assignment in agentic RL.}
Prior methods assign credit via trajectory preferences in
ETO~\citep{song2024eto}, tree-search action preferences in
Agent~Q~\citep{putta2024agentq}, utterance values in
ArCHer~\citep{zhou2024archer}, and turn-level critics in
SWEET-RL~\citep{zhou2025sweetrl}. Critic-free methods compare rollout returns
in GRPO~\citep{shao2024deepseekmath} or shared-state actions in
GiGPO~\citep{feng2025groupingroup}, while GraphGPO merges shared states into a batch graph for goal-distance
credit assignment~\citep{cheng2026graphgpo}. G-Skill persists each task graph
across rounds, turning its paths and branches into evolving skills.

\section{Conclusion}\label{sec:conclusion}

Our findings show that reusable skills for agentic RL need not be isolated
textual summaries distilled from single trajectories. By preserving structural
relations across rollouts in persistent task graphs, G-Skill recombines
rollouts into task-level procedures and refines state-level action experience
without LLM generation. Across ALFWorld and WebShop, it improves GRPO and
GraphGPO on two models with broader task coverage and closer references.
Moreover, persistent graphs adapt across horizons, driving path recombination
on long tasks and local refinement on shorter ones. Experience graphs thus
offer a lightweight alternative to LLM distillation and motivate pairing graph
memory with learned state abstractions in richer environments.

\clearpage
\section*{Limitations}

G-Skill requires a state representation that recognizes repeated states across
rollouts. When few states recur, the graph degenerates into separate
trajectories with few shared nodes. Conversely, an overly coarse
representation can alias distinct states and yield non-executable paths.

Even a closely related skill can mislead the policy during transfer: it may
mistake a reference detail for a current fact or apply a mismatched
constraint, causing wrong searches or invalid choices.

Because every goal-reaching state yields a Step Skill, bank growth increases
storage and index-maintenance overhead, and retrieval must distinguish useful
references from an expanding pool of similar skills at each step.

Finally, our evaluation covers two structured environments and 3B--4B models,
evaluates ALFWorld on \texttt{valid\_seen}, and reports sampling variance
rather than independent training seeds. Testing robustness across training
seeds, larger models, and less structured domains remains future work.


\bibliography{custom}

\clearpage
\appendix
\onecolumn
\section{Notation Summary}\label{app:notation}

Table~\ref{tab:notation} summarizes all mathematical symbols defined across
our task formulation, persistent task graphs, dual-granularity skill
construction, skill evolution, and retrieval components.

\begin{table}[h!]
\centering
\footnotesize
\renewcommand{\arraystretch}{1.12}
\begin{tabular}{@{}l p{0.64\textwidth} l@{}}
\toprule
\textbf{Symbol} & \textbf{Description} & \textbf{Ref.} \\
\midrule
\multicolumn{3}{@{}l}{\textit{Agentic Task Formulation and Policy Learning}} \\
$\mathcal{D},\, x,\, I$ & Task distribution, task instance $x\sim\mathcal{D}$, and textual task description of $x$ & Sec.~\ref{sec:preliminary} \\
$t,\, T$ & Interaction step index ($t\in\{1,\ldots,T\}$) and episode horizon & Sec.~\ref{sec:preliminary} \\
$s_t,\, a_t,\, h_t$ & Current state, selected action, and interaction history ($I$, $s_t$, and recent turns) at step $t$ & Sec.~\ref{sec:preliminary} \\
$\pi_\theta,\, J(\theta)$ & Agent policy parameterized by $\theta$ and expected-return RL objective & Sec.~\ref{sec:preliminary} \\
$\tau,\, r(\tau)$ & Episode trajectory $(s_1,a_1,\ldots,s_T,a_T)$ and final task reward & Sec.~\ref{sec:preliminary} \\
$\mathcal{B},\, \bar{\mathcal{B}},\, z$ & Skill bank, frozen snapshot used during a rollout round, and an individual skill $z\in\mathcal{B}$ & Sec.~\ref{sec:preliminary}, App.~\ref{app:algorithm} \\
$\mathcal{R},\, \mathcal{Z}_t$ & Retrieval operator and retrieved skill set $\mathcal{Z}_t=\mathcal{R}(h_t,\mathcal{B})\subseteq\mathcal{B}$ at step $t$ & Sec.~\ref{sec:preliminary}, \ref{sec:rl-loop} \\
$r,\, R,\, \mathcal{X}$ & Training round index ($1,\ldots,R$), total training rounds, and sampled task batch & App.~\ref{app:algorithm} \\
$N,\, \mathcal{T}_x,\, \tau_x^j$ & Rollout group size per task and rollout group $\mathcal{T}_x=\{\tau_x^j\}_{j=1}^{N}$ for task $x$ & Sec.~\ref{sec:rl-loop}, App.~\ref{app:algorithm} \\
\midrule
\multicolumn{3}{@{}l}{\textit{Persistent Task Graphs and Skill Evolution}} \\
$\mathcal{G}=(\mathcal{S},\mathcal{E}),\, \mathcal{G}_x$ & Directed state-transition graph (for task $x$) with node set $\mathcal{S}$ and edge set $\mathcal{E}$ & Sec.~\ref{sec:traj-agg}, App.~\ref{app:algorithm} \\
$s_1,\, s^*$ & Shared initial state across rollouts of a task and terminal goal node for success & Sec.~\ref{sec:traj-agg} \\
$(s,a,s')$ & Observed transition edge from state $s$ to successor state $s'$ via environment-normalized action $a$ & Sec.~\ref{sec:traj-agg}, \ref{sec:skill-const} \\
$n,\, \mathcal{G}^{(n)},\, \Delta\mathcal{G}^{(n)}$ & Task sampling round index, graph after round $n$, and round-$n$ transition increment ($\mathcal{G}^{(0)}=\varnothing$) & Sec.~\ref{sec:traj-agg} \\
$d^{(n)}(s),\, C^{(n)}(s)$ & Updated round-$n$ goal distance and ranked Step Skill action list & Sec.~\ref{sec:skill-evol} \\
\midrule
\multicolumn{3}{@{}l}{\textit{Graph-Grounded Skill Construction}} \\
$\mathcal{P}(s),\, |p|$ & Set of directed paths from state $s$ to goal $s^*$, and number of transitions in path $p$ & Sec.~\ref{sec:skill-const} \\
$d(s)$ & Shortest graph distance from state $s$ to goal $s^*$ ($+\infty$ if unreachable) & Sec.~\ref{sec:skill-const} \\
$p,\, z_{\mathrm{task}}$ & Selected shortest known path from $s_1$ to $s^*$ with $|p|=d(s_1)$, and Task Skill $z_{\mathrm{task}}=(I,p)$ & Sec.~\ref{sec:skill-const} \\
$\mathcal{A}(s)$ & Eligible actions at state $s$ with at least one observed successor state $s'$ satisfying $d(s')<+\infty$ & Sec.~\ref{sec:skill-const} \\
$H(s,a)$ & Action helpfulness, defined as $\max_{(s,a,s')\in\mathcal{E},\,d(s')<+\infty}[d(s)-d(s')]$ & Sec.~\ref{sec:skill-const} \\
$K,\, C(s),\, z_{\mathrm{step}}(s)$ & Action cap per Step Skill ($K=3$), Top-$K$ action--score list, and Step Skill $z_{\mathrm{step}}(s)=(I,s,C(s))$ & Sec.~\ref{sec:skill-const}, App.~\ref{app:training-config} \\
\midrule
\multicolumn{3}{@{}l}{\textit{Skill Retrieval and Hyperparameters}} \\
$I_r,\, s_r$ & Reference task description and reference state of a candidate skill & App.~\ref{app:retrieval-impl} \\
$\operatorname{sim}_{\mathrm{task}},\, \operatorname{sim}_{\mathrm{step}}$ & Task Skill similarity $\rho(I,I_r)$ and weighted Step Skill similarity over task and state & App.~\ref{app:retrieval-impl} \\
$\lambda$ & Weight on task similarity in Step Skill retrieval ($\lambda=0.5$, state weight $1-\lambda$) & App.~\ref{app:retrieval-impl}, \ref{app:training-config} \\
$q,\, v,\, v',\, |v|,\, \bar{\ell}$ & BM25 query text, candidate text, candidate index variable, candidate length, and average document length & App.~\ref{app:retrieval-impl} \\
$\mathcal{U}(q),\, f(w,v),\, \operatorname{IDF}(w)$ & Unique tokens in query $q$, frequency of token $w$ in candidate $v$, and inverse document frequency & App.~\ref{app:retrieval-impl} \\
$k_1,\, b,\, L(v)$ & BM25 saturation ($k_1=1.5$) and length-normalization ($b=0.75$) parameters, and length factor & App.~\ref{app:retrieval-impl}, \ref{app:training-config} \\
$B(q,v),\, \rho(q,v)$ & Raw BM25 score and normalized BM25 similarity scaled by query self-score and best candidate & App.~\ref{app:retrieval-impl} \\
$\gamma$ & Distance discount factor used in GraphGPO baselines ($0.1$ on ALFWorld, $0.2$ on WebShop) & App.~\ref{app:training-config} \\
\bottomrule
\end{tabular}
\caption{Summary of mathematical notation used throughout the paper.}
\label{tab:notation}
\end{table}
\clearpage
\twocolumn

\section{G-Skill Algorithm and Implementation}\label{app:implementation}
\subsection{Training Procedure}\label{app:algorithm}

Algorithm~\ref{alg:gskill} summarizes the interaction between policy learning
and skill evolution. Here, $\mathcal{D}$ is the task distribution and
$\mathcal{X}$ is a batch sampled from it. The policy $\pi_\theta$ is optimized
for $R$ training rounds indexed by $r\in\{1,\ldots,R\}$. For each task $x$, $\mathcal{G}_x$ denotes its
persistent task graph, and
$\mathcal{T}_x=\{\tau_x^j\}_{j=1}^{N}$ denotes the group of $N$ rollouts
collected in one training round. The skill bank $\mathcal{B}$ contains the
Task and Step Skills derived from all task graphs. Its snapshot
$\bar{\mathcal{B}}$ is held fixed during rollout collection in a training
round, so every rollout in the group uses the same version of the skills.

\FloatBarrier
\begin{algorithm}[t]
\caption{G-Skill training procedure}
\label{alg:gskill}
\begin{algorithmic}[1]
\REQUIRE Policy $\pi_\theta$, task distribution $\mathcal{D}$, training rounds $R$, rollout group size $N$
\STATE Initialize $\mathcal{G}_x\leftarrow\varnothing$ for each task $x$
\STATE Initialize $\mathcal{B}\leftarrow\varnothing$
\FOR{training round $r=1,\ldots,R$}
  \STATE Sample a task batch $\mathcal{X}$ from $\mathcal{D}$
  \STATE Take a skill-bank snapshot $\bar{\mathcal{B}}\leftarrow\mathcal{B}$
  \FOR{each task $x\in\mathcal{X}$}
    \STATE Collect $\mathcal{T}_x=\{\tau_x^j\}_{j=1}^{N}$ using $\pi_\theta$ and $\bar{\mathcal{B}}$
  \ENDFOR
  \FOR{each task $x\in\mathcal{X}$}
    \STATE Merge transitions from $\mathcal{T}_x$ into $\mathcal{G}_x$
    \STATE Recompute distances and update Task and Step Skills in $\mathcal{B}$
  \ENDFOR
  \STATE Compute rewards and advantages for the collected rollouts
  \STATE Update $\theta$ using the selected base RL objective
\ENDFOR
\end{algorithmic}
\end{algorithm}

\subsection{Environment-Specific State Representations}\label{app:state-rep}

\paragraph{ALFWorld.}
Each task instance is identified by its game file. Following the state
representation used by GraphGPO, a graph node is keyed by the raw observation,
the agent's current location and held item, the recorded item-movement history,
and the sorted set of admissible commands. Two states are considered identical
when all of these fields match. The command set is used only to
distinguish graph nodes, and it is omitted from the observation shown in a
retrieved skill. Informational actions such as \texttt{examine}, \texttt{look},
\texttt{inventory}, and \texttt{help}, as well as actions returning
\texttt{Nothing happens}, are treated as state-preserving transitions and
therefore form self-loops. Ordinary edges require admissible actions. The final successful action
always leads to $s^*$.

\paragraph{WebShop.}
Each shopping instruction defines a task instance. The environment wrapper
strips a \texttt{[SEP]}-delimited instruction prefix when present. Under
WebShop's standard text observation format used in our experiments, the
formatted page text, together with its instruction header, is retained verbatim as
both the graph-node identity and the skill observation. Identical formatted
pages are treated as the same state within a task graph. A transition that leaves the page
representation unchanged is retained as a self-loop. After
\texttt{click[search]} produces a self-loop, the same action is removed from
the immediately following action pool to prevent repeated ineffective clicks.
Search and click actions are otherwise validated against the actions available
on the current page.

State identities are local to a task instance in both environments. Thus,
identical state strings from different tasks are stored in different graphs
and are compared only during skill retrieval.

\subsection{Graph and Skill Implementation}\label{app:graph-skill-impl}

\paragraph{Graph construction and persistence.}
For every training trajectory, G-Skill retains admissible nonterminal
transitions from both successful and unsuccessful rollouts. The final active
transition of a successful rollout is redirected to the shared goal node
$s^*$. Duplicate edges are stored once, while self-loops are preserved. After a
training rollout group is completed, its transitions are merged into the
corresponding task graphs. Distances are then recomputed by a breadth-first
traversal of the reverse graph starting at $s^*$. Each task graph is stored
separately and reloaded when training resumes.

\paragraph{Task Skills.}
All rollouts of the same task share an identical initial state, which is used
as the starting node for Task Skill extraction. A Task Skill is materialized
by following edges that reduce the distance to $s^*$ by one.
When several such edges are available, ties are resolved deterministically by
the action string and then the successor-state key. The resulting path is
serialized as an alternating observation--action history ending in
\texttt{SUCCESS}.
The stored Task Skill payload contains the reference task description and this
history.

\paragraph{Step Skills.}
During graph construction, ALFWorld actions are normalized by removing entity
indices so the policy treats them as transferable patterns rather than literal
commands. WebShop actions are normalized only by lowercasing and whitespace
normalization. Multiple observed transitions from the same state may therefore
share one normalized action label while leading to different successor states.
For each non-goal state connected to $s^*$, G-Skill computes the distance
reduction separately for each eligible transition. Preserved self-loop
transitions, including informational actions such as \texttt{look}, actions
returning \texttt{Nothing happens}, and page-preserving WebShop clicks, are
not excluded from ranking and contribute a distance-reduction score of $0$
before aggregation over normalized actions. Each normalized action receives
the maximum helpfulness across its eligible transitions, as defined in
Equation~\ref{eq:helpfulness}, so every goal-reaching state has at least one
action with helpfulness $1$. Normalized actions are ranked by this score,
breaking ties deterministically by the action string, and the Step Skill
exposes up to three actions by default. The policy is instructed to
instantiate each pattern using the actions available in its current state.
Appendix~\ref{app:skill-examples} shows the resulting Task and Step Skills for
both environments.

\subsection{Skill Retrieval and Context Injection}\label{app:retrieval-impl}

G-Skill builds separate BM25 indexes for task descriptions and state
representations. Text is lowercased and tokenized into alphanumeric tokens,
with term-saturation parameter $k_1=1.5$ and length-normalization parameter
$b=0.75$. Let $\mathcal{U}(q)$ denote the unique tokens in query $q$, $f(w,v)$
the frequency of token $w$ in candidate $v$, $|v|$ the token length of $v$,
$\operatorname{IDF}(w)$ the inverse document frequency of $w$, and
$\bar{\ell}$ the average document length. With length-normalization
factor $L(v)$, the raw BM25 score $B(q,v)$ is
\begin{equation}
 \begin{aligned}
  L(v)&=1-b+b\frac{|v|}{\bar{\ell}},\\
  B(q,v)&=\sum_{w\in\mathcal{U}(q)}
  \operatorname{IDF}(w)
  \frac{f(w,v)(k_1+1)}{f(w,v)+k_1 L(v)}.
 \end{aligned}
\end{equation}
We temporarily add the query to the index when computing $\operatorname{IDF}$
and $\bar{\ell}$. Its self-score $B(q,q)$ and the highest score
across all candidates $v'$ set the scale for normalized similarity $\rho(q,v)$:
\begin{equation}
 \rho(q,v)=
 \frac{B(q,v)}{\max\{B(q,q),\max_{v'}B(q,v')\}},
\end{equation}
with zero score when the denominator is zero. For ALFWorld, we remove entity indices from the task and state texts used
for retrieval. The displayed observation retains the original indices.

Candidates are partitioned by task type: ALFWorld uses its six task families,
whereas WebShop uses one shopping category. We exclude candidates with
identical task descriptions. Given $I$, $s_t$, and a candidate's reference
description $I_r$ and reference state $s_r$, Task Skill similarity
$\operatorname{sim}_{\mathrm{task}}$ and Step Skill similarity
$\operatorname{sim}_{\mathrm{step}}$ with task weight $\lambda\in[0,1]$ are
\begin{equation}
 \begin{aligned}
  \operatorname{sim}_{\mathrm{task}}(I,I_r)
  &=\rho(I,I_r),\\
  \operatorname{sim}_{\mathrm{step}}(I,s_t;I_r,s_r)
  &=\lambda\rho(I,I_r)\\
  &\quad +(1-\lambda)\rho(s_t,s_r).
 \end{aligned}
\end{equation}
The default configuration selects up to one Task Skill and two Step Skills
with $\lambda=0.5$ and applies no minimum similarity threshold.

Retrieved skills are serialized with fixed templates and inserted before the
current task description. The templates identify them as reference experience
and require the policy to ground every action in the current state and action
pool. The complete templates are provided in
Appendix~\ref{app:prompt-templates}. If the resulting prompt exceeds the
configured prompt budget, both retrieved skill blocks are omitted for that
interaction. When an action forms a self-loop, the previous Step Skill is
reused at the next interaction because the underlying state has not changed.

\section{Experimental Details}\label{app:experimental-details}

\subsection{Environments}\label{app:environments}

\paragraph{ALFWorld.}
ALFWorld~\citep{shridhar2021alfworld} aligns embodied household tasks from
ALFRED~\citep{shridhar2020alfred} with interactive text games powered by
TextWorld~\citep{cote2018textworld}. An agent explores rooms, opens
receptacles, manipulates household objects, and applies state changes to
satisfy a goal across six task families: Pick, Clean, Cool, Look, Heat,
and Pick2. At each turn, the agent receives a textual observation of its
current room or receptacle together with the list of admissible commands,
and executes one command within a 50-step horizon. Rewards are sparse and
binary, granting a positive return only upon completing the goal, while
invalid actions incur a small penalty. Evaluation covers all 140 runnable
games in the \texttt{valid\_seen} split, reporting per-family and overall
success rates over five seeds.

\paragraph{WebShop.}
WebShop~\citep{yao2022webshop} is a simulated e-commerce environment in
which an agent locates and purchases a product satisfying a natural-language
shopping instruction specifying item type, attributes, options such as color
or size, and a price cap. At each turn within a 15-step horizon, the agent
observes the formatted text page and chooses either a search query or a
clickable page button such as inspecting an item, selecting an option,
returning to results, or clicking \texttt{Buy Now}. Upon purchase, the
environment computes a continuous task score in $[0,1]$ reflecting product,
attribute, option, and price alignment, and marks an episode as successful
when the score reaches $1$. Evaluation uses the same 64 benchmark slots for
every method and reports average task score and success rate.

\subsection{Baselines}\label{app:baselines}

\paragraph{Base Model.}
Base Model evaluates the shared initialization checkpoint for each model
and benchmark without RL updates. For Qwen3-4B-Instruct-2507 on WebShop,
this initialization is the checkpoint fine-tuned for one epoch on SkillRL's
public WebShop demonstrations using the \texttt{qwen3\_nothink} template in
LLaMA-Factory~\citep{zheng2024llamafactory}.

\paragraph{GRPO.}
GRPO~\citep{shao2024deepseekmath} is a critic-free policy optimization
baseline that samples a group of $N$ rollouts per task and normalizes
episode returns within each group to estimate advantages without using
external skills.

\paragraph{GraphGPO.}
GraphGPO~\citep{cheng2026graphgpo} aggregates rollouts for the same task
within a single training batch into a state-transition graph and derives
step-level advantages from shortest distances to the goal. Unlike G-Skill,
its graph is discarded after each batch rather than forming a persistent
cross-task skill bank across rounds.

\paragraph{SkillRL.}
SkillRL~\citep{xia2026skillrl} distills trajectories with a teacher LLM
into a hierarchical library of general strategies, task-specific advice,
and common mistakes. During GRPO training, it retrieves relevant skills
and periodically prompts the teacher to add new principles from failed
rollouts.

\paragraph{D2Skill.}
D2Skill~\citep{tu2026d2skill} maintains a dynamic dual-granularity skill
bank storing task-level plans and step-level reflections generated by a
teacher LLM. It estimates skill utility by comparing rollouts with and
without skills and prunes whole skills falling below a threshold.

\paragraph{Skill1.}
Skill1~\citep{shi2026skill1} trains a single unified policy via RL to
rerank retrieved skills, condition actions on selected skills, and distill
new skills from task outcomes without relying on an external teacher
model.

\paragraph{Reproduction details.}
We implement G-Skill and run GRPO and GraphGPO in the same verl-based
framework~\citep{sheng2025hybridflow}. GraphGPO retains its official
per-batch reward graph and advantage estimator, while G-Skill maintains its
persistent skill graphs separately. SkillRL, D2Skill, and Skill1 are run
from their released implementations under the same model initialization,
training budget, and evaluation protocol whenever supported. We
use DeepSeek-V4.1-Flash-1017~\citep{deepseekai2026deepseekv41flash} to
generate skills for SkillRL and D2Skill. Other settings follow each
baseline's official configuration.

\subsection{Evaluation Protocol and Training Configuration}\label{app:training-config}

For evaluation, we sample at temperature 0.4 with environment seed 0 and
sampling seeds $\{0,1,2,3,4\}$, reporting the mean and standard deviation
across these five runs of the same policy. During evaluation, all skill-augmented methods use their final frozen skill banks
without test-time updates. G-Skill builds its bank exclusively from training
rollouts, while baselines retain their prescribed initialization and update
protocols. Each ALFWorld game runs once per sampling
seed, and training-time validation metrics serve only as diagnostics rather
than standardized test results.

Training runs for 140 steps with 16 tasks per batch and eight rollouts per
task, yielding 128 trajectories before filtering. Each run uses four NVIDIA
H100 80GB GPUs, vLLM for generation~\citep{kwon2023vllm}, and SwanLab for
logging. Tables~\ref{tab:training-hparams}--\ref{tab:gskill-hparams} list the
shared optimization, environment, and retrieval settings.

\begin{table*}[t]
\centering
\small
\begin{tabular}{llr}
\toprule
\textbf{Category} & \textbf{Hyperparameter} & \textbf{Value} \\
\midrule
\multirow{9}{*}{Optimization}
& Training steps & 140 \\
& Learning rate & $1\times10^{-6}$ \\
& PPO mini-batch size & 64 \\
& PPO epochs per update & 1 \\
& PPO clip range & 0.2 \\
& Entropy coefficient & 0.001 \\
& Gradient clipping & 1.0 \\
& Weight decay & 0.01 \\
& Learning-rate warmup ratio & 0 \\
\midrule
\multirow{3}{*}{Regularization}
& KL-loss coefficient & 0.01 \\
& KL-loss type & Low-variance KL \\
& KL penalty in reward & Disabled \\
\midrule
\multirow{8}{*}{Rollout and validation}
& Task batch size & 16 \\
& Rollouts per task ($N$) & 8 \\
& Rollout batch size & 128 \\
& Maximum prompt length & 8,192 \\
& Maximum response length & 512 \\
& Training temperature & 1.0 \\
& Top-$p$ / Top-$k$ & 1.0 / $-1$ \\
& Validation batch size / interval & 64 / every 5 steps \\
\midrule
\multirow{4}{*}{System}
& Numerical precision & BFloat16 \\
& Gradient checkpointing & Enabled \\
& Number of GPUs & 4 \\
& vLLM tensor parallelism & 2 \\
\bottomrule
\end{tabular}
\caption{Shared training hyperparameters.}
\label{tab:training-hparams}
\end{table*}

\begin{table}[t]
\centering
\small
\begin{tabular}{@{}l@{\hspace{8pt}}c@{\hspace{8pt}}c@{}}
\toprule
\textbf{Hyperparameter} & \textbf{ALFWorld} & \textbf{WebShop} \\
\midrule
History length & 2 & 2 \\
Environment seed & 0 & 0 \\
Maximum interactions & 50 & 15 \\
PPO micro-batch size per GPU & 8 & 2 \\
Invalid-action penalty coefficient & 0.1 & 0.1 \\
GraphGPO distance discount $\gamma$ & 0.1 & 0.2 \\
vLLM GPU memory utilization & 0.55 & 0.50 \\
\bottomrule
\end{tabular}
\caption{Training settings by environment.}
\label{tab:environment-hparams}
\end{table}

\begin{table}[t]
\centering
\small
\begin{tabular}{lr}
\toprule
\textbf{Hyperparameter} & \textbf{Value} \\
\midrule
Retrieval method & BM25 \\
BM25 $k_1$ & 1.5 \\
BM25 $b$ & 0.75 \\
Task Skills retrieved & 1 \\
Step Skills retrieved & 2 \\
Actions per Step Skill ($K$) & 3 \\
Task-similarity weight ($\lambda$) & 0.5 \\
State-similarity weight ($1-\lambda$) & 0.5 \\
\bottomrule
\end{tabular}
\caption{G-Skill retrieval hyperparameters.}
\label{tab:gskill-hparams}
\end{table}

\subsection{Ablation Configurations}\label{app:ablation-configs}

We use Qwen2.5-3B-Instruct with GRPO on WebShop and evaluate each variant
with the same five sampling seeds. Table~\ref{tab:ablation-configs} lists
the changes to G-Skill for each variant in Table~\ref{tab:skill_ablation}.

\begin{table}[!ht]
\centering
\small
\begin{tabular}{@{}l@{\hspace{6pt}}p{0.58\columnwidth}@{}}
\toprule
\textbf{Variant} & \textbf{Change from G-Skill} \\
\midrule
Single-Trajectory Skill & \raggedright Use one successful rollout as the only skill, in Task Skill format. \tabularnewline
w/o Task Skill & \raggedright Retrieve and inject only Step Skills. \tabularnewline
w/o Step Skill & \raggedright Retrieve and inject only Task Skills. \tabularnewline
w/o Graph Evolution & \raggedright Freeze each task graph after its first rollout group. \tabularnewline
w/ Random Retrieval & \raggedright Shuffle eligible candidates for each skill type before selection. \tabularnewline
\bottomrule
\end{tabular}
\caption{WebShop ablation configurations.}
\label{tab:ablation-configs}
\end{table}

\subsection{Reference Task Comparison}\label{app:reference-map}

We compare the source tasks of Task Skills in the D2Skill, Skill1, and
G-Skill banks. For Skill1, we reconstruct the evaluation banks from the
skills retained at evaluation. We encode the source and test descriptions
with Qwen3-Embedding-0.6B~\citep{zhang2025qwenembedding}. Figure~\ref{fig:reference-task-distribution}
plots distinct descriptions on the first two principal components, with
the same axes for all three banks within each environment.

For Figures~\ref{fig:nearest-reference} and~\ref{fig:nearest-reference-alfworld}
and Table~\ref{tab:paired-references}, we compare descriptions by cosine
similarity before projection. For each test task, we exclude identical
descriptions and take the closest remaining match in each bank. Each
ALFWorld game counts once, even when several games share a description.
Within each environment, we illustrate the task whose three scores have
the smallest total absolute deviation from their respective bank medians.

\setcounter{topnumber}{1}
\section{Additional Analyses}\label{app:analyses}
\subsection{Analysis of Skill-Based Baselines}\label{app:baseline-analysis}

Under the shared training budget in Table~\ref{tab:main_results}, LLM-generated skill baselines occasionally trail vanilla GRPO, most visibly on WebShop and Qwen2.5-3B ALFWorld. Inspecting saved banks and logs reveals three failure modes.

\paragraph{Skill1: formatting and distillation collapse on WebShop.}
Skill1 trains one policy to solve tasks, rerank candidates, and distill skills. On ALFWorld, both models keep a 100\% rerank parse rate and build nearly full banks of 5,000 skills for Qwen2.5-3B and 4,997 for Qwen3-4B, reaching 82.9\% on Qwen2.5-3B. On WebShop, however, structured self-generation collapses:
\begin{itemize}
\setlength{\itemsep}{1pt}\setlength{\parskip}{0pt}\setlength{\topsep}{2pt}
\item On Qwen3-4B WebShop, success drops to 10.6\% with \texttt{0.0\%} rerank parse and distillation rates across all five seeds and 64 test tasks, leaving \texttt{0} skills and pure prompt overhead.
\item On Qwen2.5-3B WebShop, success reaches 37.8\% as the rerank parse rate falls to \texttt{1.6\%--3.1\%} across seeds and only 49 skills survive by the end of training, limiting coverage.
\end{itemize}

\paragraph{D2Skill: high eviction churn from whole-skill pruning.}
D2Skill uses a teacher LLM to reflect on failures and prunes whole skills with negative estimated utility. During Qwen2.5-3B training, 18 pruning rounds on ALFWorld evict \textbf{502} skills (247 task, 255 step). On WebShop, 21 rounds evict \textbf{950} skills (459 task, 491 step), leaving only 309+309 and 310+311 skills. Both generic rules and narrow tips face wholesale eviction:
\begin{itemize}
\setlength{\itemsep}{1pt}\setlength{\parskip}{0pt}\setlength{\topsep}{2pt}
\item On ALFWorld, \texttt{step\_072}, ``\emph{Use the visible mug at the coffeemachine instead of searching elsewhere}'', created at step 12, was retrieved 316 times across unrelated steps before eviction at utility $-0.94$. Similarly, \texttt{step\_067}, ``\emph{Don't re-examine without new information}'', was retrieved 480 times with utility $-0.95$ before removal.
\item On WebShop, \texttt{task\_009}, ``\emph{Match distinctive attributes before paging}'', and \texttt{task\_115}, ``\emph{Match all constraints before purchasing}'', were both evicted at utility $-0.38$, showing how noisy whole-skill attribution drives rapid churn.
\end{itemize}

\paragraph{SkillRL: abstract principles and early bank saturation.}
SkillRL initializes from offline teacher principles, 55 on ALFWorld and 66 on WebShop, and adds principles during RL. By the end of Qwen2.5-3B training, its banks stay small at 127 ALFWorld principles over 24 rounds and 93 WebShop principles over 9 rounds, with no updates after step 45. Its skills remain abstract heuristics, such as \texttt{gen\_001} on WebShop, ``\emph{Include the product type, 1--2 key functional attributes, and any hard constraints (price, size, color) in the search query...}'', and \texttt{pic\_001} on ALFWorld, ``\emph{Maintain a checklist of all visible and closed containers/surfaces...}''. For 3B--4B policies, such advice is harder to ground into exact actions than G-Skill's concrete paths and scored actions.

\subsection{ALFWorld Reference Similarity and Pairwise Bank Comparisons}\label{app:alfworld-reference}

\begin{figure}[!t]
    \centering
    \includegraphics[width=\columnwidth]{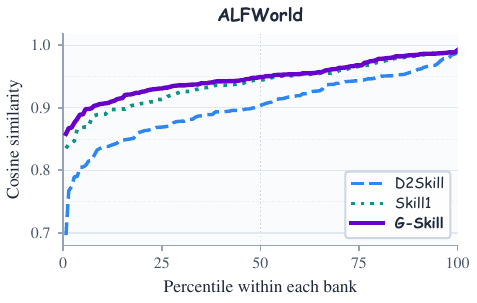}
    \caption{\textbf{Closest non-identical reference similarity on ALFWorld with Qwen2.5-3B.}
    For all 140 test task descriptions, we compute Qwen3-Embedding-0.6B cosine similarity against non-identical source tasks in each bank. Curves plot each task's highest score by percentile (x-axis), where higher curves show closer matches.}
    \label{fig:nearest-reference-alfworld}
\end{figure}

Figure~\ref{fig:nearest-reference-alfworld} shows the ALFWorld nearest-reference curve complementing Figure~\ref{fig:nearest-reference}, and Table~\ref{tab:nearest-reference-alfworld} illustrates a typical cooling game whose similarities across the three banks lie closest to their medians. Because ALFWorld has a structured template space, G-Skill and Skill1 track closely across percentiles even though Skill1 stores 5,000 Task Skills versus G-Skill's 1,351.

\begin{table}[!ht]
\centering
\footnotesize
\begin{tabular*}{\columnwidth}{@{\extracolsep{\fill}}llll@{}}
\toprule
\textbf{Source} & \textbf{Operation} & \textbf{Object} & \textbf{Destination} \\
\midrule
\textbf{Test task} & \textbf{Cool} & \textbf{Plate} & \textbf{Cabinet} \\
\midrule
D2Skill & \textbf{Cool} & Bowl & \textbf{Cabinet} \\
Skill1  & \textbf{Cool} & \textbf{Plate} & Shelf \\
\textbf{G-Skill} & \textbf{Cool} & \textbf{Plate} & Shelf \\
\bottomrule
\end{tabular*}
\caption{\textbf{Nearest stored reference tasks for a typical ALFWorld cooling test game.}
Selected as the test game whose nearest-reference scores across the three banks lie closest to their medians. Each row shows a bank's closest non-identical source task by cosine similarity, with matched attributes in \textbf{bold}.}
\label{tab:nearest-reference-alfworld}
\end{table}

Table~\ref{tab:paired-references} directly compares each pair of banks on the \emph{same} test instance. On WebShop, G-Skill gives a strictly closer match than Skill1 on 100.0\% and D2Skill on 70.3\% of requests. On ALFWorld, it beats D2Skill on 73.6\% and matches or exceeds Skill1 on 96.4\%, sharing the closest task on 70.0\% of games.

\begin{table}[!ht]
\centering
\small
\begin{tabular*}{\columnwidth}{@{}l@{\extracolsep{\fill}}lrrr@{}}
\toprule
& & \multicolumn{3}{c}{Nearest reference (\%)} \\
\cmidrule(l){3-5}
Environment & Other bank & G-Skill & Same & Other \\
\midrule
ALFWorld & D2Skill & 73.6 & 25.0 & 1.4 \\
         & Skill1  & 26.4 & 70.0 & 3.6 \\
\addlinespace[2pt]
WebShop  & D2Skill & 70.3 & 20.3 & 9.4 \\
         & Skill1  & 100.0 & 0.0 & 0.0 \\
\bottomrule
\end{tabular*}
\caption{\textbf{Pairwise comparison of closest available reference descriptions per test task.}
For each test task description, we compare its highest Qwen3-Embedding-0.6B cosine similarity against non-identical stored source tasks in G-Skill versus D2Skill or Skill1.
Columns report the share of test tasks where \textbf{G-Skill} is strictly closer, both banks share the \textbf{Same} closest source task, or the \textbf{Other} bank is strictly closer.}
\label{tab:paired-references}
\end{table}

\subsection{Failures in Skill Transfer}\label{app:skill-transfer-failures}

\begin{figure}[!ht]
    \centering
    \includegraphics[width=.94\columnwidth]{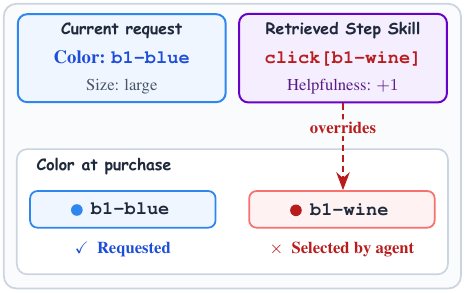}
    \caption{\textbf{Buying the reference color on WebShop.} A positively scored retrieved Step Skill action (\texttt{click[b1-wine]}, helpfulness $+1$) overrides the shopper's requested color (\texttt{b1-blue}).}
    \label{fig:transfer-failures}
\end{figure}

\paragraph{Reference constraints replace current goals.}
In two Qwen3-4B validation episodes with retrieved Step Skills, Figure~\ref{fig:transfer-failures} shows a WebShop color choice where the reference and current requests share size, yet the shopper asks for \texttt{b1-blue}. Although both colors appear on the page, the agent selects \texttt{b1-wine}, matching the positively scored action in the retrieved Step Skill.

\begin{figure}[!ht]
    \centering
    \includegraphics[width=.88\columnwidth]{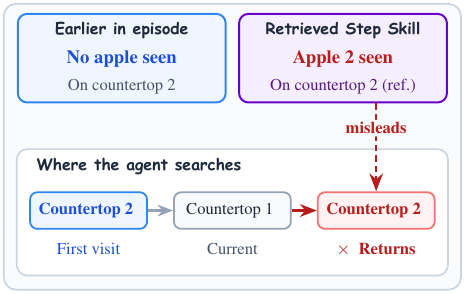}
    \caption{\textbf{Revisiting a countertop in ALFWorld.} An apple seen in a retrieved reference episode (ref.) misleads the agent into re-checking an empty countertop.}
    \label{fig:transfer-failure-alfworld}
\end{figure}

\paragraph{Reference facts become current beliefs.}
In Figure~\ref{fig:transfer-failure-alfworld}, the agent has visited
countertops 2 and 1 without seeing an apple. It returns to countertop 2 after mistaking the reference apple for
one it saw there. The search ends without success.

\subsection{Displayed Action Comparisons and Aggregation Breakdown}
\label{app:skill-updates}

\begin{figure}[!t]
    \centering
    \includegraphics[width=.92\columnwidth]{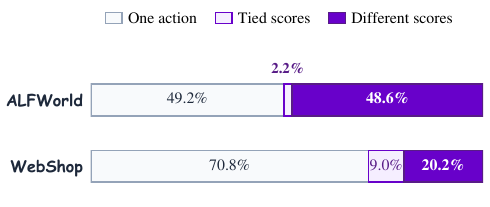}
    \caption{\textbf{Action diversity inside evaluation Step Skills on Qwen2.5-3B with GRPO.}
    Share of Step Skills showing one action, tied helpfulness scores, or distinct helpfulness scores.}
    \label{fig:action-comparisons}
\end{figure}

Figure~\ref{fig:action-comparisons} shows evaluation Step Skill action lists. WebShop shows 70.8\% single-action skills and 20.2\% distinct-score comparisons due to shorter episodes, whereas ALFWorld balances single-action (49.2\%) and distinct-score skills (48.6\%).

In Table~\ref{tab:skill-evolution}, ALFWorld shortcut rates peak on multi-stage families at 48.1\% (Pick2), 42.6\% (Cool), and 40.2\% (Heat), followed by 25.9\% (Clean), 22.1\% (Look), and 10.4\% (Pick). A skill is eligible for revision if its task is resampled after creation, counting shorter paths for Task Skills and Top-3 changes for Step Skills.

\section{Prompts and Skill Examples}\label{app:prompts-and-examples}
\subsection{Prompt Templates}\label{app:prompt-templates}

Figure~\ref{fig:policy-templates} shows where the policy receives retrieved skills and the current task context. Highlighted placeholders mark the fields filled at each step.

\begin{figure*}[p]
\input{sections/appendix/prompts_and_examples/prompt_alfworld}
\input{sections/appendix/prompts_and_examples/prompt_webshop}
\caption{Skill instructions and policy prompts for ALFWorld and WebShop.}
\label{fig:policy-templates}
\end{figure*}

\subsection{Skill Examples}\label{app:skill-examples}

Figure~\ref{fig:skill-examples} reproduces one Task Skill and one Step
Skill recorded during training in each environment.

\begin{figure*}[p]
\setlength{\smallskipamount}{2pt}
\begin{gskilltaskbox}{Task Skill Example (ALFWorld)}\footnotesize\raggedright
\textbf{task:} examine the alarmclock with the desklamp.

\smallskip
\textbf{history:}
You are in the middle of a room. Looking quickly around you, you see a bed 1,
a desk 1, a drawer 5, a drawer 4, a drawer 3, a drawer 2, a drawer 1, a
dresser 1,\newline
and a garbagecan 1. Location: middle of a room. Items in hand
(status): nothing.

\smallskip
\hspace*{1em}\texttt{--go to desk-->}\par
You arrive at desk 1. On the desk 1, you see a alarmclock 4, a alarmclock 3,
a alarmclock 2, a alarmclock 1, a book 1, a cd 2, a cd 1, a creditcard 3, a
creditcard 2, a desklamp 1, a keychain 3,\newline
a pen 1, and a pencil 3. Location:
desk 1. Items in hand (status): nothing.

\smallskip
\hspace*{1em}\texttt{--take alarmclock from desk-->}\par
You pick up the alarmclock 1 from the desk 1. Location: desk 1. Items in hand
(status): alarmclock 1(unprocessed).

\smallskip
\hspace*{1em}\texttt{--use desklamp--> SUCCESS}
\end{gskilltaskbox}

\begin{gskillstepbox}{Step Skill Example (ALFWorld)}\footnotesize\raggedright
\textbf{task:} put some lettuce on countertop.

\smallskip
\textbf{observation:}
You open the fridge 1. The fridge 1 is open. In it, you see a apple 2, a apple
1, a egg 1, a lettuce 1, a plate 2, and a potato 3. Location: fridge 1. Items
in hand (status): nothing.

\smallskip
\textbf{actions:}\par
\hspace*{1em}\textbf{action:} take lettuce from fridge\quad \textbf{helpfulness:} 1

\smallskip
\hspace*{1em}\textbf{action:} go to countertop\quad \textbf{helpfulness:} -1
\end{gskillstepbox}

\begin{gskilltaskbox}{Task Skill Example (WebShop)}\footnotesize\raggedright
\textbf{task:} Find me mirrors with rose gold for hair removal, and price lower
than 40.00 dollars

\smallskip
\textbf{history:}
WebShop Instruction: Find me mirrors with rose gold\newline
for hair removal, and price lower than 40.00 dollars [button] Search [button\_]

\smallskip
\hspace*{1em}\texttt{--search[mirrors rose gold hair removal price <40]-->}\par
Instruction: Find me mirrors with rose gold for hair removal, and price lower
than 40.00 dollars [button] Back to Search [button\_] Page 1 (Total results:
50) [button] Next > [button\_] [button] B084VWH12H [button\_] Look Bella 10x
Magnifying Compact Folding Double Mirror. Magnification for Hair Removal.
Perfect for Purses. Distortion Free - Rose Gold \$11.99 [button] B08ZKDQPZ3
[button\_] Hair Tinsel Kit Strands With Tool 47 Inch 12 Colors 2100 Strands
Fairy Hair Tinsel Kit Hair Extensions Sparkling Glitter Shiny Silk Tinsel (12
colors) \$13.96 [button] B079Y4B4YP [button\_] INOAR PROFESSIONAL - Smoothing
System - Deep Cleansing Shampoo \& Treatment Set (8.45 oz x 2) \$70.0 [button]
B08QCL5SX5 [button\_] XIAOQIAO Wall Mirror Decorative Antique Mirror, Wall
Hanging Metal Frame Nordic Style Art, for Decor Hotel Hallway Mantel Living
Room \$395.14 [button] B00FR4V65M [button\_] Polder Styling Station, White
\$21.54 [button] B085GLK7X4 [button\_] Slip Silk Sleep Mask, Hollywood Hills
(One Size) - 100\% Pure Mulberry 22 Momme Silk Eye Mask - Comfortable Sleeping
Mask with Elastic Band + Pure Silk Filler and Internal Liner \$89.99 [button]
B08PFRP4RN [button\_] 10 Packs Hair Storage Carrying Bags Big Satin Drawstring
Pouches Gift Bag for Hair Extentions, Bundles, Wigs, Hair Tools (Rose red)
\$17.99 [button] B00T9U3NA6 [button\_] Hair Removal Pads 4 Piece Set (2-Pack)
\$4.99 [button] B08X2PKKB2 [button\_] 1 oz Aluminum Tin Jar with Screw Cap
Refillable Container for Cosmetic, Lip Balm, Cream, Rose Gold 12 Pcs. \$7.99
[button] B09LLRSWF4 [button\_] Yheakne Metal Hair Slide Clip Barrette Geometric
Hair Holder Clip Pin Vintage Hair Slide Pin Bun Holder\newline
Alloy Hair Clip
Decorative Hair Accessories for Women and Girls (Gold) \$8.5

\smallskip
\hspace*{1em}\texttt{--click[b084vwh12h]-->}\par
Instruction: Find me mirrors with rose gold for hair removal, and price lower
than 40.00 dollars [button] Back to Search [button\_] [button] < Prev
[button\_] Look Bella 10x Magnifying Compact Folding Double Mirror.
Magnification for Hair Removal. Perfect for Purses. Distortion Free - Rose Gold
Price: \$11.99 Rating: N.A. [button] Description [button\_] [button] Features
[button\_] [button] Reviews [button\_] [button] Buy Now [button\_]

\smallskip
\hspace*{1em}\texttt{--click[buy now]--> SUCCESS}
\end{gskilltaskbox}

\begin{gskillstepbox}{Step Skill Example (WebShop)}\footnotesize\raggedright
\textbf{task:} Find me 1080p hd, high definition projectors, and price lower
than 150.00 dollars

\smallskip
\textbf{observation:}
WebShop Instruction: Find me 1080p hd,\newline
high definition projectors, and price lower than 150.00 dollars [button] Search [button\_]

\smallskip
\textbf{actions:}\par
\hspace*{1em}\textbf{action:}
search[1080p hd, high definition projectors price <150.00]\quad \textbf{helpfulness:} 1

\smallskip
\hspace*{1em}\textbf{action:}
search[1080p hd, high definition projectors price < 150.00 dollars]\quad \textbf{helpfulness:} 1

\smallskip
\hspace*{1em}\textbf{action:} search[1080p hd projectors]\quad \textbf{helpfulness:} -2
\end{gskillstepbox}
\caption{Task and Step Skills recorded during training on ALFWorld and WebShop.}
\label{fig:skill-examples}
\end{figure*}

\end{document}